%% file: main.tex
\documentclass[10pt,twocolumn,letterpaper]{article}

\usepackage{cvpr}              

\usepackage{xargs}
\usepackage{booktabs} 
\usepackage{multirow}
\usepackage{colortbl}
\usepackage{comment}
\usepackage{bigints}

\usepackage[colorinlistoftodos,prependcaption,textsize=tiny]{todonotes}
\newcommandx{\change}[2][1=]{\todo[linecolor=red,backgroundcolor=red!25,bordercolor=red,#1]{#2}}
\newcommandx{\info}[2][1=]{\todo[linecolor=OliveGreen,backgroundcolor=OliveGreen!25,bordercolor=OliveGreen,#1]{#2}}
\newcommandx{\improvement}[2][1=]{\todo[linecolor=Plum,backgroundcolor=Plum!25,bordercolor=Plum,#1]{#2}}
\newcommandx{\thiswillnotshow}[2][1=]{\todo[disable,#1]{#2}}

\definecolor{red}{rgb}{1, 0.7, 0.7} 
\definecolor{orange}{rgb}{1, 0.85, 0.7} 
\definecolor{yellow}{rgb}{1, 1, 0.7} 

\definecolor{tabfirst}{rgb}{1, 0.7, 0.7} 
\definecolor{tabsecond}{rgb}{1, 0.85, 0.7} 
\definecolor{tabthird}{rgb}{1, 1, 0.7} 

\usepackage{marginnote}

\usepackage{lipsum}
\input{macros.tex}


\definecolor{cvprblue}{rgb}{0.21,0.49,0.74}
\usepackage[pagebackref,breaklinks,colorlinks,citecolor=cvprblue]{hyperref}

\def\paperID{3774} 
\def\confName{CVPR}
\def\confYear{2024}

\title{Neural Inverse Rendering from Propagating Light\vspace{-0.5em}}

\author{
  Anagh Malik$^{*1,2}$
  \space\space
  Benjamin Attal$^{*3}$
\space\space
  Andrew Xie$^{1,2}$
  \space\space
  Matthew O'Toole$^{+3}$
  \space\space
  David B. Lindell$^{+1,2}$
  \\
  \space\space
  \small{\textnormal{$^{1}$University of Toronto\space\space$^{2}$Vector Institute\space\space$^{3}$Carnegie Mellon University}}\\
    \small{$^{*}$joint first authors\space\space $^{+}$equal contribution} \\
    \small{\url{https://anaghmalik.com/InvProp}}
}

\begin{document}

\twocolumn[{%
\renewcommand\twocolumn[1][]{#1}%
\maketitle
\begin{center}
    \vspace{-2.5em}
    \captionsetup{type=figure}
        \includegraphics[width=\textwidth]{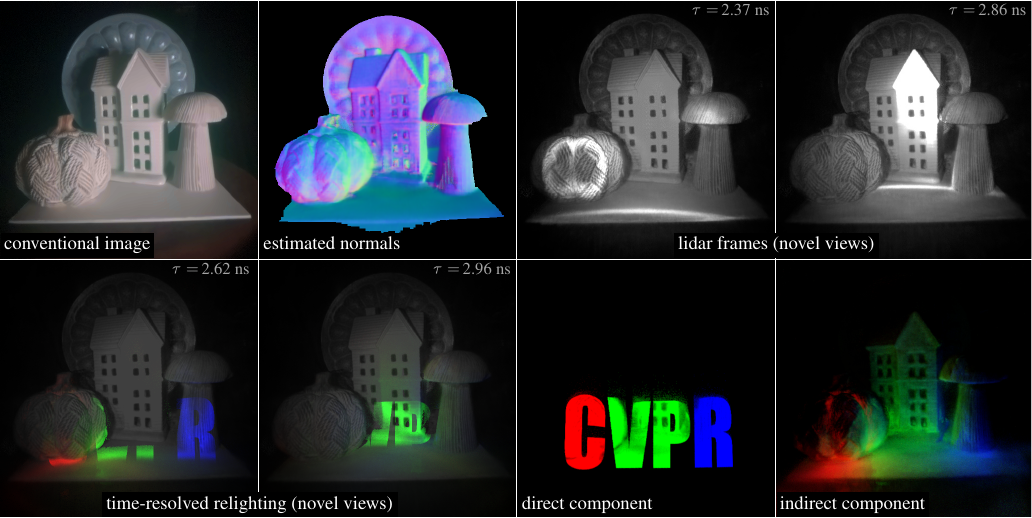}
    \\
    \vspace{-0.5em}
    \captionof{figure}{We introduce a method to model and invert multi-view, time-resolved measurements of propagating light from a flash lidar system. 
    \textbf{(row 1)}  Our method accurately recovers the geometry of this scene and enables rendering of time-resolved lidar measurements that reveal light propagation from novel views.  (\textbf{row 2}) Physically-based modeling enables novel applications, such as time-resolved relighting and automatic decomposition of light transport into direct and indirect components.}
    \label{fig:teaser}
\end{center}
}]

\maketitle

\input{sec/abstract}    
\input{sec/introduction}

\input{sec/related_work}

\input{sec/method}
\input{sec/results}
\input{sec/discussion}
\input{sec/acknowledgements}

{
    \small
    \bibliographystyle{ieeenat_fullname}
    \bibliography{main}
}
\clearpage

\appendix
\counterwithin{figure}{section}
\counterwithin{table}{section}
\input{sec/appendix}


\end{document}

%% file: macros.tex
\usepackage{xcolor}

\newcommand{\normal}{\mathbf{n}}

\newcommand{\light}{\ell}
\providecommand{\xpos}{\mathbf{x}}
\newcommand{\xposof}[1]{\mathbf{x}\lft(#1\rgt)}
\providecommand{\xorigin}{\mathbf{o}}

\providecommand{\normal}{\mathbf{n}}

\providecommand{\direction}{\boldsymbol{\omega}}
\providecommand{\dirin}{\direction_\mathrm{i}}

\providecommand{\dirout}{\direction_\mathrm{o}}

\newcommand{\featuregridapp}{\mathcal{H}^\textit{app}}
\newcommand{\featuregridgeom}{\mathcal{N}^\textit{geom}}
\newcommand{\featureapp}{\mathbf{f}^\textit{app}}

\newcommand{\brdfdir}{f^{\textit{dir}}}
\newcommand{\brdfind}{f_{\Omega}^{\textit{indir}}}
\newcommand{\radianceindir}{L_\text{i}^{\textit{dir}}}
\newcommand{\radianceinind}{L_{\text{i}, \Omega}^{\textit{indir}}}
\newcommand{\radianceinmeas}{L_\text{i}^{\textit{meas}}}

\providecommand{\ncount}{\mathrm{N}}

\providecommand{\radiancein}{L_\mathrm{i}}
\providecommand{\radianceout}{L_\mathrm{o}}
\providecommand{\cacheoutdir}{L_\mathrm{o}^{\textit{cache,dir}}}
\providecommand{\cacheoutind}{L_\mathrm{o}^{\textit{cache,indir}}}
\providecommand{\cachein}{L_\mathrm{i}^{\textit{cache}}}
\providecommand{\cacheout}{L_\mathrm{o}^{\textit{cache}}}

\newcommand\lft{\mathopen{}\left}
\newcommand\rgt{\aftergroup\mathclose\aftergroup{\aftergroup}\right}

%% file: sec/abstract.tex
\begin{abstract}
We present the first system for physically based, neural inverse rendering from multi-viewpoint videos of propagating light. Our approach relies on a time-resolved extension of neural radiance caching --- a technique that accelerates inverse rendering by storing infinite-bounce radiance arriving at any point from any direction. 
The resulting model accurately accounts for direct and indirect light transport effects and, when applied to captured measurements from a flash lidar system, enables state-of-the-art 3D reconstruction in the presence of strong indirect light. 
Further, we demonstrate view synthesis of propagating light, automatic decomposition of captured measurements into direct and indirect components, as well as novel capabilities such as multi-view time-resolved relighting of captured scenes. 
\end{abstract}

%% file: sec/introduction.tex
\section{Introduction}
\label{sec:intro}
Ultrafast imaging systems such as lidar illuminate a scene with a pulse of light and capture the backscattered ``echoes'' from the propagating wavefront~\cite{schwarz2010mapping}.
Precisely measuring the speed-of-light time delay of backscattered light enables 3D reconstruction, and so lidar systems based on this principle are popular in applications from autonomous driving~\cite{li2020lidar} to augmented reality~\cite{apple_vision_pro} and remote sensing~\cite{dong2017lidar}.

Lidar relies on time-resolved measurements of \textit{direct} light transport, or light that reflects directly from a surface back to the sensor. 
Measurements of \textit{indirect} light transport (i.e., light that scatters multiple times before reaching the sensor) are typically ignored or discarded because modeling indirect light requires computationally expensive (and often intractable) inverse rendering using procedures such as recursive path tracing~\cite{pharr2023physically,jakob2022dr}. 
Still, measurements of indirect light are a rich source of information about material properties, appearance, and geometry~\cite{naik2011single,wu2014decomposing,zhang2022modeling,o2018confocal,klinghoffer2024platonerf}.
Our work seeks to model and invert multi-viewpoint, time-resolved measurements of propagating light from a lidar system to recover scene geometry and to render videos of propagating light from novel viewpoints and under novel illumination conditions (see Figure~\ref{fig:teaser}).

Conventional lidar systems pre-process captured time-resolved measurements into a 3D point cloud, which represents an estimate of scene geometry based on direct light transport.
While recent work leverages lidar measurements captured from multiple viewpoints to perform 3D reconstruction and novel view synthesis, existing methods use a point cloud representation~\cite{rematas2022urban,deng2022depth,zhang2024nerf,huang2023neural} or direct-only time-resolved measurements~\cite{malik2023transient,luo2024transientangelo,ramazzina2024gated, mu2024towards}, and thus ignore indirect light. 
Other methods for multi-viewpoint reconstruction with active imaging use time-of-flight sensors~\cite{attal2021torf,okunev2024flowed}, or structured light~\cite{shandilya2023neural,ichimaru2024neural}, but similarly fail to explicitly model indirect light. 
Our work is close to that of Malik et al.~\cite{malik2024flying}, which uses lidar measurements and a representation based on neural radiance fields (NeRFs)~\cite{mildenhall2021nerf} to render videos of propagating light from novel viewpoints. 
However, while their representation is effective for view synthesis, it does not use a physically-based model, and so cannot reconstruct accurate geometry or render the scene under novel illumination conditions. 

Accurately modeling indirect light transport requires solving the rendering equation~\cite{kajiya1986rendering}.
Many renderers use path tracing algorithms based on Monte Carlo sampling, and differentiable versions of these methods have shown promise for inverse rendering of conventional (i.e., steady state) images~\cite{li2018differentiable,loubet2019reparameterizing,nimier2019mitsuba} and time-resolved measurements~\cite{wu2021differentiable,yi2021differentiable}, such as those from a lidar system.
However, while these methods work in constrained problem settings, such as non-line-of-sight imaging~\cite{tsai2019beyond,iseringhausen2020non}, they are difficult to deploy on captured multi-viewpoint experiments due to their computational expense and sensitivity to noise and local minima.
To address this issue, recent methods approximate the rendering equation in a hybrid fashion using a volumetric representation of geometry and physics-based appearance models~\cite{zhang2021nerfactor,srinivasan2021nerv}. 
This type of approach has proven especially effective for inverse rendering of appearance, geometry, and material parameters from conventional intensity images~\cite{zhang2022modeling,jin2023tensoir,liu2023nero,attal2024flash}.
Still, no previous technique performs inverse rendering of time-resolved direct and indirect light transport in the same fashion.

Here, we propose a method for inverse rendering from multi-viewpoint, time-resolved measurements of propagating light from a \emph{flash lidar} system --- a type of lidar that flood-illuminates the entire scene with a pulse of light.
Our approach is based on a hybrid neural representation, where we model geometry using volume rendering, and appearance using a physically-based model that simulates global illumination effects using a radiance cache~\cite{ward1988ray}.
Specifically, instead of integrating light paths using path tracing, our radiance cache stores a representation of time-resolved radiance arriving at any point in a volume from any direction.
The representation is optimized in an amortized fashion, thereby removing the need to evaluate the rendering integral recursively; instead, we need only render direct lighting from the lidar and query the radiance cache to evaluate indirect light at any surface point.
Our model builds on previous neural rendering frameworks for multi-viewpoint, time-resolved rendering of the direct~\cite{malik2023transient} and indirect components~\cite{malik2024flying} of propagating light, and grounds them in a framework for physically-based rendering.
Overall, we make the following contributions.
\begin{itemize}
    \item We propose a method for neural inverse rendering of propagating light using a physically-based model with a time-resolved radiance cache. 
    \item We capture a new dataset of multi-viewpoint, time-resolved flash lidar measurements with calibrated light source and camera positions.
    \item We demonstrate our method in simulation and on captured data;  we show state-of-the-art results in geometry reconstruction under strong indirect light transport, and we render videos of propagating light from novel viewpoints and under novel illumination conditions in scenes with varying reflectance properties and significant indirect light transport effects.  
\end{itemize}

%% file: sec/related_work.tex
\section{Related Work}
\label{sec:related_work}
Our approach brings together the areas of time-resolved imaging and rendering, as well as inverse rendering. 

\vspace{-0.5em} \paragraph{Time-resolved imaging.}
Time-of-flight systems such as lidar measure the time of flight by marking the arrival time of a backscattered pulse of light~\cite{koechner1968optical}. 
These systems usually combine nanosecond or picosecond pulsed lasers with fast photodiodes~\cite{huntington2020ingaas} or single-photon avalanche diodes (SPADs)~\cite{rapp2020advances,kirmani2014first, shin2015photon} to measure ultrafast variations in incident light. 
The resulting time-resolved measurements capture direct and indirect light transport, and can be used to record videos of propagating light at ultrafast timescales~\cite{velten2013femto,o2014temporal,gariepy2015single,lindell2018towards}.   
While continuous-wave time-of-flight systems~\cite{li2014time,gupta2015phasor} can also be used to capture light propagation, their temporal resolution is more limited (e.g., nanoseconds rather than picoseconds)~\cite{heide2013low,o2014temporal}.

Our approach uses a lidar system with a picosecond pulsed laser and a SPAD to capture multi-view videos of propagating light. 
Similar to previous work~\cite{chen2019learning,o2018confocal,rapp2020seeing,xin2019theory,faccio2020non,malik2023transient,malik2024flying}, we repeatedly illuminate the scene with pulses of light from the laser. 
The SPAD detects the arrival times of individual photons with picosecond-level accuracy and outputs a photon count histogram that approximates the time-resolved waveform of incident light~\cite{o2017reconstructing}.
Our approach is the first to capture a multi-view dataset of photon count histograms where both the flash lidar light source and sensor vary in position.
Moreover, we develop the first technique for physically-based time-resolved inverse rendering using multi-viewpoint videos of propagating light.

\vspace{-0.5em} \paragraph{Time-resolved rendering.}
Time-resolved path-tracing renderers simulate wavefronts of propagating light~\cite{jarabo2014framework} and account for effects such as birefringence~\cite{nimier2019mitsuba}, refraction~\cite{pediredla2020path}, and volumetric scattering~\cite{jarabo2018bidirectional,pediredla2019ellipsoidal}.
Recently, differentiable versions of these time-resolved renderers~\cite{yi2021differentiable,wu2021differentiable} have been developed; however, robust analysis-by-synthesis scene reconstruction using these methods is an open problem due to computational complexity and sensitivity to initialization and noise. 
Other renderers approximate the time-resolved light transport matrix~\cite{o2014temporal,ramesh20085d} for specific imaging problems such as non-line-of-sight imaging.
In this area, techniques have been developed to model two-bounce~\cite{tsai2016shape,klinghoffer2024platonerf}, three-bounce~\cite{liu2019non,lindell2019wave,xin2019theory,rapp2020seeing,seidel2023non}, or higher-order scattering events~\cite{lindell2020three,royo2023virtual} to recover occluded geometry. 
In the non-line-of-sight setting, renderers usually assume that light reflects off of planar surfaces with known (usually diffuse) reflectance properties, or else model specific scattering paths rather than arbitrary light transport~\cite{faccio2020non}. Although several works consider indirect light transport in the context of continuous-wave time-of-flight sensors, they usually treat it as a residual to be removed when estimating depth~\cite{gupta2015phasor, naik2015light, adam2016bayesian}.
We perform physically-based modeling of multiply scattered light without restrictive assumptions on scene geometry or material properties and integrate our approach into an inverse rendering framework for scene reconstruction from captured measurements.

\vspace{-0.5em} \paragraph{Physically-based inverse rendering.}
Inverse rendering aims to recover scene attributes, like materials, lighting, and geometry, from a set of images~\cite{ramamoorthi2001signal,yu1999inverse,debevec1998rendering}. 
Existing techniques use a physically-based model of light transport~\cite{pharr2023physically,veach1998robust,kajiya1986rendering} and differentiable rendering with gradient-based optimization to decompose the scene's appearance into its constituent attributes~\cite{chen2019learning,li2018differentiable,jakob2022dr}. 
Recent techniques for physically-based rendering using NeRFs~\cite{mildenhall2021nerf} have made inverse rendering considerably more robust, but either only consider direct illumination~\cite{srinivasan2021nerv,zhang2021nerfactor,verbin2024eclipse,boss2021neuralpil,boss2022samurai}, or require explicitly simulating multiple light bounces to model indirect light~\cite{mai2023neural}, which is computationally expensive.
Another approach is to use radiance caches --- data structures that store the hemisphere of incoming radiance at every point~\cite{ward1988ray}. 
This hemisphere is then integrated against the local bidirectional reflectance distribution function (BRDF)~\cite{pharr2023physically} to yield outgoing radiance, which is optimized to match the observed image pixels.  
Combining the radiance cache with NeRFs leads to more efficient modeling of indirect light~\cite{yao2022neilf,liu2023nero,zhang2022modeling,attal2024flash, wu2023nefii}.

Finally, we note that the concurrent work of Wu et al.~\cite{wu2024gani} makes use of radiance-caching-based inverse rendering with a collocated point light source and color camera. 
However, no previous technique performs physically-based inverse rendering from multi-viewpoint time-resolved measurements.
To this end, we develop a new, time-resolved radiance cache, enabling neural inverse rendering from videos of propagating light.

%% file: sec/method.tex
\begin{figure*}[th!]
    \vspace{-1em}
    \begin{center}
    \includegraphics[width=\textwidth]{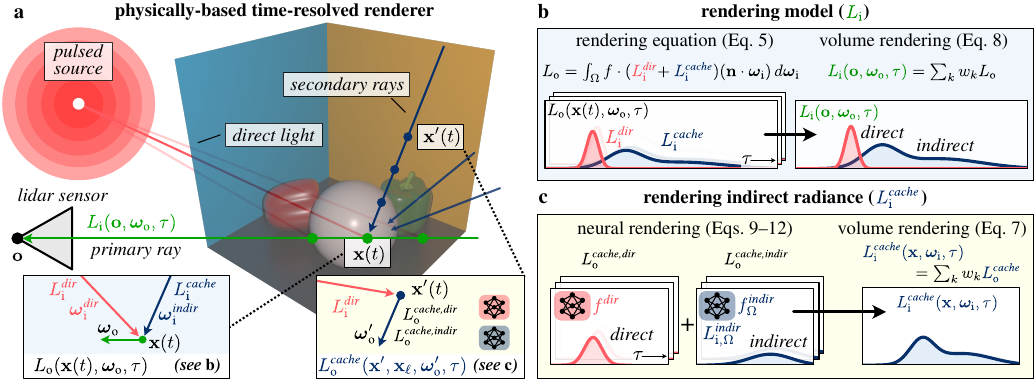}
    \end{center}
    \vspace{-1.5em}
    \caption{Method overview.
    (\textbf{a}) Our time-resolved renderer combines physically-based rendering for primary rays \textit{(left inset)}, and neural rendering for an indirect radiance cache along secondary rays \textit{(right inset)}. 
    (\textbf{b}) The incident radiance $\radiancein$ at a sensor pixel is a function of the outgoing radiance $\radianceout$ from each point $\xposof{t}$ along a sensor ray, which integrates incident direct light $\radianceindir$ and indirect light $\cachein$ from a pulsed laser source. 
    (\textbf{b}, \textit{left}) The rendering equation computes the outgoing radiance as an integral over the positive hemisphere $\Omega$ (with respect to the normal $\normal$) of the incident radiance. (\textbf{b}, \textit{right}) Applying volume rendering to the outgoing radiance yields the incident radiance at a sensor pixel.
(\textbf{c}) The radiance cache is used to evaluate indirect radiance $\cachein$ by casting secondary rays and querying neural networks at points $\xpos'(t)$. (\textbf{c}, \textit{left}) The networks output the reflectance $\brdfdir$ of direct light $\radianceindir$, indirect reflectance $\brdfind$, and indirect incident radiance $\radianceinind$. These quantities are used to calculate the time-resolved direct ($\cacheoutdir$) and indirect  ($\cacheoutind$) outgoing radiance in the direction $\dirout'$.
(\textbf{c}, \textit{right}) Volume rendering the outgoing radiance along a secondary ray yields $\cachein(\xpos, \dirin, \tau)$.
We optimize the scene appearance and geometry to enforce consistency between rendered and captured measurements.
}

    \vspace{-1.25em}
    \label{fig:method}
\end{figure*}

\section{Background: Radiance Caching with NeRFs}
\label{sec:radiance-caching}

The rendering equation~\cite{kajiya1986rendering}, models the outgoing radiance of light in direction $\dirout$ at a point $\xpos$ along a ray $\xposof{t} = \xorigin - t\dirout$ with origin $\xorigin$ and ray parameter $t$: 
\begin{align}
\resizebox{0.895\columnwidth}{!}{%
$\radianceout(\xposof{t}, \dirout) =  \bigintsss_{\mathrm{\Omega}} f\lft(\xpos, \dirin, \dirout\rgt) \radiancein\lft(\xpos, \dirin\rgt) \lft(\mathbf{n} \cdot \dirin\rgt)\, d \dirin\,$}.
    \label{eqn:rendering-equation}
\end{align}
The equation integrates the incident radiance $\radiancein$ arriving to $\xpos$ from direction $\dirin$ weighted by the BRDF $f$. 
The integral is over the positive hemisphere with respect to the normal $\normal$: $\mathrm{\Omega} = \{\dirin : \mathbf{n}\cdot \dirin > 0 \}$.
 Naive evaluation of the rendering equation leads to an exponential increase in computation since the equation must be evaluated recursively to compute the incident radiance.

To avoid this computational penalty, we leverage radiance caching, which removes the problematic recursion by replacing incident radiance $\radiancein$ in the rendering equation with a look-up into a cache $\cachein$:
\begin{align}
\resizebox{0.895\columnwidth}{!}{%
$\radianceout(\xposof{t}, \dirout) =  \bigintsss_{\mathrm{\Omega}} f\lft(\xpos, \dirin, \dirout\rgt) \cachein\lft(\xpos, \dirin\rgt) \lft(\mathbf{n} \cdot \dirin\rgt)\, d \dirin\,$}.
    \label{eqn:rendering-equation-cache}
\end{align}
The integral can be efficiently approximated, e.g., by sampling the cache and the BRDF using multiple importance sampling~\cite{pharr2023physically}. 

Recent work demonstrates that NeRFs provide accurate modeling of the radiance cache~\cite{jin2023tensoir,ling2024nerf,attal2024flash}. 
Specifically, we can compute $\cachein\lft(\xpos, \dirin\rgt)$ by volume rendering the NeRF along a secondary ray  $\xpos'(t) = \xorigin' - t\dirout'$, where $\xorigin' = \xpos$ and $\dirout' = \dirin$~\cite{drebin1988volume,lombardi2019neural,mildenhall2021nerf}:
\begin{align}
    \cachein(\xorigin', \dirin) &= \sum_{k = 1}^{\ncount} w_k \,
    \cacheout\lft(\xpos'(t_k), \dirout' \rgt). \label{eqn:quadrature}
\end{align}
Here, $\cacheout$ is the outgoing radiance at each point along the secondary ray predicted by the NeRF. 
The values $w_k$ are quadrature weights that account for the transmittance and absorption along the ray, calculated as a function of the density $\sigma$ at each sample point $\xpos'(t_k)$ and the ray interval $(\Delta t)_k$~\cite{max1995optical,tagliasacchi2022volume}:
\begin{align}
\resizebox{0.895\columnwidth}{!}{%
$w_k = \lft(1 - e^{-\sigma(\xpos'(t_k)) (\Delta t)_k} \rgt)e^{\lft( - \sum_{j=1}^{k - 1} \sigma(\xpos'(t_j))(\Delta t)_j\rgt)}\,$}.
\end{align}

Our work adapts this observation to time-resolved rendering based on lidar measurements.

\section{Method}
We model and invert time-resolved light transport, including direct and indirect effects, to recover scene geometry and material properties. 
Our method uses a physically-based time-resolved renderer with a time-resolved radiance cache parameterized by neural networks, as shown in Figure~\ref{fig:method}.
We perform inverse rendering by optimizing the representation using measurements of propagating light from a flash lidar system.

\subsection{Physically-Based Time-Resolved Rendering}
We model a lidar measurement by casting a primary ray $\xposof{t} = \xorigin -t\dirout$ into the scene. 
For each point along that ray, we render the outgoing radiance, $\radianceout$, in the direction of the sensor. 
Our time-resolved rendering equation is a modified version of Equation~\ref{eqn:rendering-equation}, where we add the time of flight $\tau$ as 
\begin{align}
    &\radianceout(\xposof{t}, \dirout, \tau) \nonumber \\
    &=  \int_{\mathrm{\Omega}} f\lft(\xposof{t}, \dirin, \dirout\rgt) \radiancein\lft(\xposof{t}, \dirin, \tau\rgt) \lft(\mathbf{n} \cdot \dirin\rgt)\, d \dirin. 
    \label{eqn:tof-rendering-equation}
\end{align}
The reflectance $f$, is modeled using the Disney--GGX BRDF~\cite{burley2012physically}, which depends on the material properties of the scene as described in the appendix.

We further decompose the incident radiance into two components $\radiancein = \radianceindir + \cachein$: a direct component $\radianceindir$, and an indirect component $\cachein$ evaluated using the radiance cache.
The direct component is given as 
\vspace{-1em}
\begin{align}
    \radianceindir(\xposof{t}, \dirin, \tau)  = \frac{\delta(\direction_{\light} - \dirin)L_\text{i}^{\light}\lft(\direction_{\light}, \tau - \frac{||\xpos(t) - \xpos_{\light}||}{c}\rgt)}{||\xposof{t} - \xpos_{\light}||^2},
\label{eqn:direct}
\end{align}
\vspace{-1.25em}

\noindent where $\xpos_\light$ is the position of the light source, $\direction_\light$ is the direction from the light source,  $\radiancein^{\light}$ is the light source intensity, and $\delta(\direction_{\light} - \dirin)$ is a Dirac delta function. 
We model the inverse-square law intensity falloff and the time delay to position $\xposof{t}$ based on the speed of light $c$. 

Similar to the steady-state case (Equation~\ref{eqn:quadrature}), the time-resolved radiance cache $\cachein\lft(\xpos, \dirin, \tau\rgt)$ is evaluated using secondary rays $\xpos'(t) = \xorigin' - t\dirout'$ cast from points on the primary ray $\xorigin' = \xpos$ with $\dirout' = \dirin$.
We use time-resolved volume rendering~\cite{attal2021torf,malik2023transient,gkioulekas2016evaluation} to render the radiance cache as 
\vspace{-1.5em}
\begin{align}
    \cachein(\xorigin', \dirin, \tau) = \sum_{k = 1}^{\ncount} w_k \,
    \cacheout\lft(\xpos'(t_k), \xpos_\light, \dirout', \tau - \frac{t_k}{c} \rgt), 
    \label{eqn:tof-quadrature-cache}
\end{align}
\vspace{-0.75em}

\noindent where $\cacheout$ is the outgoing radiance predicted by a neural representation. The above states that the light incident at $\xorigin'$ in direction $\dirin$ at time $\tau$ is the sum of delayed copies of the light leaving each point $\xpos'$ along $\dirout'$, and the delay depends on the distance to $\xorigin'$ (given by the ray parameter $t_k$). 

After evaluating the time-resolved rendering equation (Equation~$\ref{eqn:tof-rendering-equation}$) for each point on the primary ray, the lidar measurement is computed by volume rendering the outgoing radiance in the same way as for the cache:
\vspace{-0.5em}
\begin{align}
    \radiancein(\xorigin, \dirout, \tau) &= \sum_{k = 1}^{\ncount} w_k \,
    \radianceout\lft(\xposof{t_k}, \dirout, \tau - \frac{t_k}{c} \rgt). \label{eqn:tof-quadrature}
\end{align}

\subsection{Time-Resolved Radiance Cache}
\paragraph{Representation.}
The cache is parameterized using a multi-resolution hash encoding $\featuregridapp$~\cite{muller2022instant,barron2023zip} to learn a position-dependent appearance feature $\featureapp$. 
Similarly, a hash encoding-based neural network $\featuregridgeom$ represents scene geometry through density and normals.
That is, 
\begin{equation}
    \featureapp = \featuregridapp(\xpos), \quad \sigma, \normal = \featuregridgeom(\xpos).
    \label{eqn:featuregrid}
\end{equation}
The density values used for volume rendering are shared across both the physically-based model and the radiance cache (i.e., Equations~\ref{eqn:tof-quadrature-cache} and~\ref{eqn:tof-quadrature}).
The appearance features are used to compute the radiance cache, which we decompose into direct and indirect components as
\begin{align}
    &\cacheout = \cacheoutdir + \cacheoutind.
    \label{eqn:tof-rendering-equation-direct-indirect}
\end{align}
We describe each of these components as follows.
\paragraph{Direct light.} The direct component is due to light that is emitted from the lidar source at $\xpos_\light$, propagates to a point $\xpos'$ in the scene, and scatters directly to the ray origin $\xorigin'$:
\begin{align}
    &\cacheoutdir(\xpos'(t), \xpos_{\light}, \dirout', \tau) \nonumber \\
    &= \brdfdir\lft(\featureapp, \normal, \direction_{\light}, \dirout'\rgt)\, \radianceindir(\xpos'(t), \direction_\light, \tau) \lft(\mathbf{n} \cdot \direction_{\light}\rgt).
    \label{eqn:tof-rendering-equation-direct}
\end{align}
Here, $\brdfdir$ is a neural network that learns the BRDF (see the appendix for a detailed description). 
In practice, we discretize the equation and compute a vector that represents radiance at each time interval.

\paragraph{Indirect light.} 
We leverage a split-sum approximation~\cite{karis2013real} to efficiently cache outgoing indirect light:
\begin{align}
    &\cacheoutind(\xpos'(t), \xpos_{\light}, \dirout', \tau) \nonumber \\
    &\approx \int_{\mathrm{\Omega}} f\lft(\xpos', \dirin', \dirout'\rgt) \lft(\mathbf{n} \cdot \dirin'\rgt)\, d \dirin' \cdot \int_{\mathrm{\Omega}} \radiancein\lft(\xpos', \xpos_{\light}, \dirin', \tau\rgt)  d \dirin' \nonumber \\
    &= \brdfind(\featureapp, \mathbf{n}, \dirout') \cdot \radianceinind(\featureapp, \xpos_{\light}, \normal, \dirout'),
    \label{eqn:tof-cache-indirect}
\end{align}
where $\brdfind$ and $\radianceinind$ are neural networks that predict the integrated BRDF and integrated incident radiance, respectively. We include conditioning on $\xpos_{\light}$, as the indirect light depends on light source position.
Following Malik et al.~\cite{malik2024flying}, $\radianceinind$ predicts a vector that represents radiance over discretized time intervals. 

\subsection{Inverse Rendering from Propagating Light}
The lidar system captures the time-resolved incident radiance at the sensor $\radianceinmeas$.
We optimize the representation by minimizing the difference between the lidar measurements and the output $\radiancein$ of the physically-based renderer (we omit their dependence on $\xorigin$, $\dirout$, and $\tau$ for brevity):
\begin{align}
    \mathcal{L}_{\textit{data}} = \sum_{\xorigin, \dirout, \tau} \alpha(\radiancein^{\textit{cache}}) \lVert\radiancein - \radianceinmeas \rVert^2.
    \label{eqn:data-loss}
\end{align}
As the cache can also be used to render the time-resolved incident radiance at the sensor, we supervise it in the same fashion by minimizing $\mathcal{L}_\textit{cache}$, which replaces $\radiancein$ with $\cachein$ in the above. 
The function $\alpha$ is chosen to more strongly penalize errors in darker regions, which improves perceptual quality similar to applying a tonemapping curve ~\cite{mildenhall2022rawnerf}:
\begin{align}
    \alpha(L) = \Big| \sum_{\tau} L \Big| ^ {-\beta},
\end{align}
where $\beta$ is a hyperparameter. Note that in Equation~\ref{eqn:data-loss}, we compute this weight using the incoming radiance from the cache ($\radiancein^{\textit{cache}}$ instead of $\radiancein$), as it is not affected by Monte Carlo render noise.

Following Hadadan et al.~\cite{hadadan2023inverse}, we leverage a radiometric prior that constrains the direct and indirect light rendered using the cache to be consistent with the full physically-based model. 
Specifically, we render the direct and indirect components of radiance at sample points $\xpos$ along the primary ray, using the cache ($\radianceout^{\textit{cache,dir/indir}}$) and the physically-based model (i.e., $\radianceout^{\textit{dir/indir}}$, given by evaluating Equation~\ref{eqn:tof-rendering-equation} using only $\radianceindir$ or $\radiancein^{\textit{cache}}$, respectively). We constrain the cache using the loss:
\begin{align}
\resizebox{1\columnwidth}{!}{%
    $\mathcal{L}_{\textit{dir/indir}} = \sum_{\xpos, \dirout, \tau} \alpha(\radianceout^{\textit{cache,dir/indir}}) \lVert \radianceout^{\textit{cache,dir/indir}} - \radianceout^{\textit{dir/indir}} \rVert^2$}.
    \label{eqn:tof-radiometric}
\end{align}

Finally, the complete photometric loss function is
\begin{align}
    \mathcal{L}_{\textit{data}} + \lambda_{\textit{cache}} \mathcal{L}_{\textit{cache}}
    + \lambda_{\textit{dir}} \mathcal{L}_{\textit{dir}}
    + \lambda_{\textit{indir}} \mathcal{L}_{\textit{indir}},
\end{align}
where $\lambda_{\textit{cache}}$, $\lambda_{\textit{dir}}$, and $\lambda_{\textit{indir}}$ are hyperparameters that weigh each loss component.
By minimizing this loss function, the method recovers a material model for the scene (parameterized using the Disney--GGX model) and the scene geometry, normals, and appearance parameters. 
In addition to the above photometric loss, we include a regularizer $\mathcal{L}_{\textit{normals}}$ that ties predicted normals to analytic normals from the density field~\cite{verbin2022refnerf}; a smoothness penalty $\mathcal{L}_{\textit{geom}}$ on the analytic normals; a smoothness penalty $\mathcal{L}_{\textit{mat}}$ on the predicted BRDF parameters; proposal resampling and distortion losses $\mathcal{L}_{\textit{interlevel}}$ and $\mathcal{L}_{\textit{distortion}}$ as in Zip-NeRF~\cite{barron2023zip}; and a mask loss $\mathcal{L}_{\textit{mask}}$.

We use multiple importance sampling for secondary rays based on the BRDF and a learnable importance sampler for incident illumination as in Attal et al.~\cite{attal2024flash}. 
The learnable importance sampler is supervised with a loss $\mathcal{L}_{\textit{vMF}}$. 
We represent the time-resolved direct outgoing light as a one-hot vector, where each bin corresponds to a discrete time interval, and indirect light as a dense vector of the same size (where the size depends on the dataset).
A complete description is provided in the appendix.

\begin{figure}[t!]
    \vspace{-5px}
    \includegraphics[width=\columnwidth]{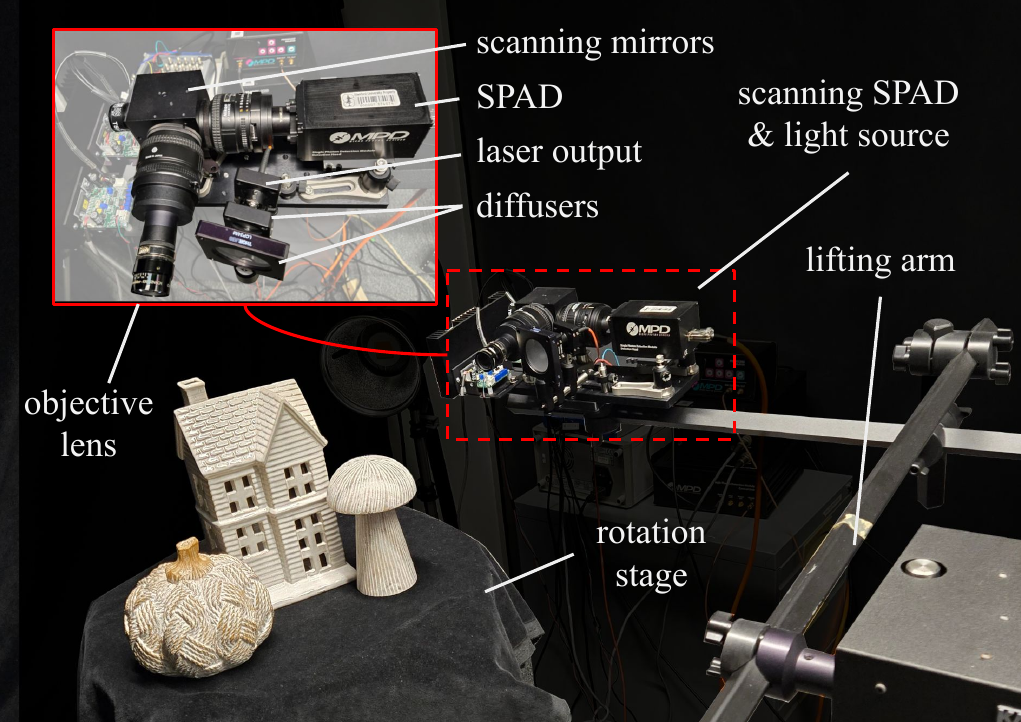}
    \caption{Multi-view capture setup. An elevation arm controls the elevation angle and a rotation stage controls the azimuth angle of the scanning SPAD. Laser light is out-coupled from an optical fiber through a collimating lens and diffusers onto the scene.}
    \label{fig:hardware}
    \vspace{-1em}
\end{figure}

\begin{figure*}[th!]
    \begin{center}
    \includegraphics[width=\textwidth]{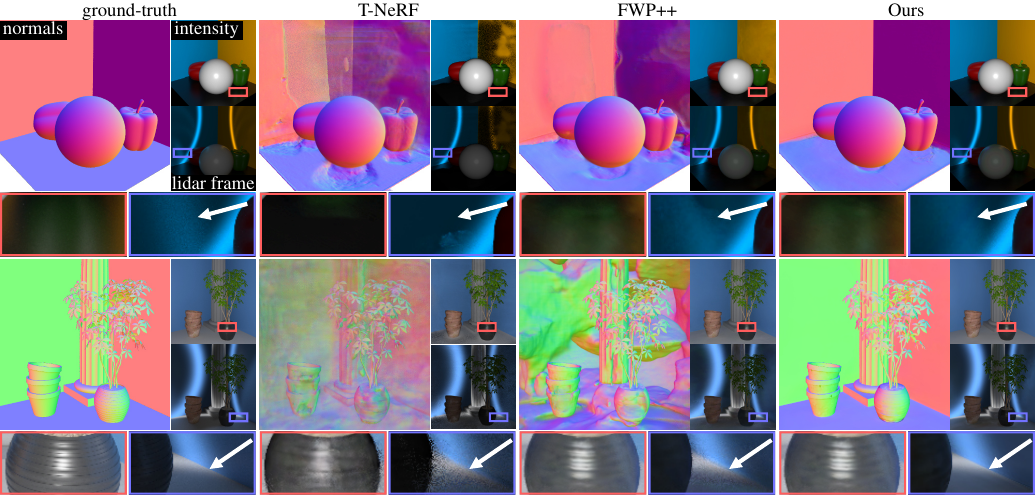}
    \end{center}
        \vspace{-15px}
    \caption{Simulated results. Compared to the baselines our method recovers more accurate normals and similar or improved intensity images due to physically-based modeling of time-resolved indirect light transport (see arrows in the lidar frame insets).}
    \vspace{-5px}
    \label{fig:simulation}
\end{figure*}

\newpage

%% file: sec/results.tex
\section{Results}

\begin{figure*}[th!]
    \begin{center}
    \includegraphics[width=\textwidth]{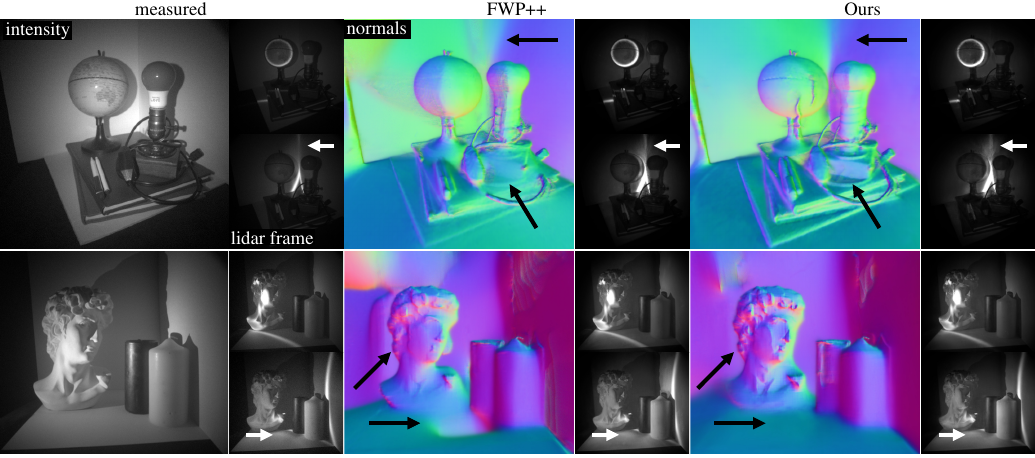}
    \end{center}
        \vspace{-15px}
    \caption{Results on the captured dataset. Our method recovers more accurate normals compared to FWP++ (cols.\ 3, 5) due to its physically-based modeling of indirect light transport effects (visible in the individual lidar frames; cols. 4, 6). Areas where FWP++ predicts incorrect normals usually correspond to regions with indirect light (arrows). }
    \vspace{-1em}
    \label{fig:capture}
\end{figure*}

\begin{figure}[h!]
    \begin{center}
    \includegraphics[width=0.48\textwidth]{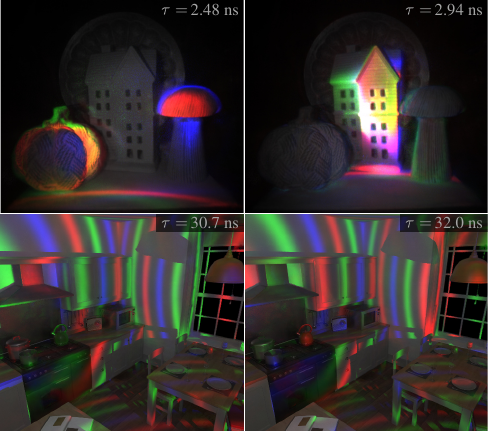}
    \end{center}
        \vspace{-15px}
        \caption{Relighting results using three different light sources (color-coded in the RGB channels) for the captured \textit{house} scene (top) and the simulated \textit{kitchen} scene (bottom).}
    \vspace{-1em}
    \label{fig:relighting}
\end{figure}

We evaluate our system on three tasks: (1) view synthesis of time-resolved lidar measurements, (2) view synthesis of integrated (steady-state) lidar images, and (3) geometry reconstruction.

\vspace{-1em} \paragraph{Evaluation metrics.} To assess the rendered integrated lidar images, we use PSNR, SSIM, and LPIPS~\cite{zhang2018unreasonable}. To evaluate the accuracy of the recovered time-resolved measurements, we use the transient intersection-over-union (transient IOU) introduced by Malik et al.~\cite{malik2024flying}.
We use mean absolute error (MAE) and L1 error to measure the accuracy of the recovered normals and depth, respectively. We report quantitative results for both simulated and captured scenes in Table~\ref{tab:results}.

\vspace{-1em} \paragraph{Baselines.} We compare our method to the following state-of-the-art time-resolved neural rendering techniques: \mbox{T-NeRF}~\cite{malik2023transient}, which uses a neural representation and a rendering model for time-resolved lidar measurements (but which only accounts for the direct component of light), and Flying with Photons (FWP)~\cite{malik2024flying}, which predicts time-resolved radiance at every point in space. Since the original version of FWP does not model light sources or movement of light sources, we use a modified version (FWP++) equivalent to our full, time-resolved radiance cache. We implement both baselines using the same hash-encoding-based neural representation~\cite{muller2022instant}, and we use the same regularizers and hyperparameters for fairness.

\subsection{Simulated Results}
\paragraph{Dataset.} We use a modification~\cite{royo2022non} of the Mitsuba 2 renderer~\cite{nimier2019mitsuba} to render multi-view transient data for three small-scale, object-centric synthetic scenes, \textit{Cornell box}, \textit{pots}, \textit{peppers}, as well as one room scale scene, \textit{kitchen}~\cite{bitterli16resources}.

\vspace{-1em} \paragraph{Comparison.} In Figure~\ref{fig:simulation}, we show integrated lidar scans from novel viewpoints, recovered normals, and individual lidar frames for the \textit{peppers} and \textit{pots} scenes. 
Since T-NeRF~\cite{malik2023transient} only models direct light, it fails to recover accurate geometry under strong indirect light from specular reflections and diffuse inter-reflections. In particular, it introduces floating artifacts to explain these effects and hence fails to predict novel views accurately. 

On the other hand, FWP++~\cite{malik2024flying} models both direct and indirect radiance and recovers accurate integrated novel views and lidar frames. 
However, because it does not use a physically-accurate rendering model, it overfits to the data. 
Specifically, it is free to use a mirror copy of the scene to explain specular reflections (e.g., on the floor of the \textit{peppers} scene or the partially specular blue walls in the \textit{pots} scene).
Further, it uses incorrect depths to explain diffuse inter-reflections (e.g., at the wall corners in the \textit{peppers} scene or along the fluting of the column in the \textit{pots} scene).

\begin{table}
    \captionof{table}{Evaluation of lidar rendering from novel viewpoints and geometry recovery. Each dataset (\textit{sim} and \textit{real}) contains 4 scenes.}
    \label{tab:results}
    \vspace{-0.5em}
    \centering
    \setlength{\tabcolsep}{2pt}
    \resizebox{\columnwidth}{!}{
    \begin{tabular}{llcccccc}
        \toprule
        & method & PSNR (dB)$\,\uparrow$ & LPIPS$\,\downarrow$ & SSIM$\,\uparrow$  & MAE$\,\downarrow$  & L1 depth$\,\downarrow$ & T-IOU$\,\uparrow$ \\\midrule
        \parbox[t]{6mm}{\multirow{3}{*}{\rotatebox[origin=c]{90}{\textit{sim}}}} & 
    T-NeRF~\cite{malik2023transient} &  \cellcolor{tabthird}22.44 &  \cellcolor{tabthird}0.40 &  \cellcolor{tabthird}0.71 &  \cellcolor{tabthird}28.00 &  \cellcolor{tabthird}0.59 &  \cellcolor{tabthird}0.58 \\
    & FWP++~\cite{malik2024flying}    & \cellcolor{tabsecond}29.00 &  \cellcolor{tabfirst}0.30 & \cellcolor{tabsecond}0.87 & \cellcolor{tabsecond}22.80 & \cellcolor{tabsecond}0.47 & \cellcolor{tabsecond}0.73 \\
    & ours   &  \cellcolor{tabfirst}30.99 & \cellcolor{tabsecond}0.31 &  \cellcolor{tabfirst}0.89 &  \cellcolor{tabfirst}8.45 &  \cellcolor{tabfirst}0.21 &  \cellcolor{tabfirst}0.76
    \\
        \toprule
        \parbox[t]{6mm}{\multirow{3}{*}{\rotatebox[origin=c]{90}{\textit{cap}}}}& 
T-NeRF~\cite{malik2023transient} &  \cellcolor{tabthird}14.67 &  \cellcolor{tabthird}0.53 &  \cellcolor{tabthird}0.35 & 
--- &  --- &  \cellcolor{tabthird}0.23 \\
& FWP++~\cite{malik2024flying}   &  \cellcolor{tabfirst}28.45 &  \cellcolor{tabfirst}0.32 &  \cellcolor{tabfirst}0.81 &  --- &  --- &  \cellcolor{tabfirst}0.55 \\
& ours   & \cellcolor{tabsecond}27.39 & \cellcolor{tabsecond}0.33 & \cellcolor{tabsecond}0.80 &  --- &  --- & \cellcolor{tabsecond}0.54        
        \\\bottomrule
    \end{tabular}}
\vspace{-1em}
\end{table}

\subsection{Captured Results}

\paragraph{Dataset.} We use a hardware setup (Figure \ref{fig:hardware}) similar to that of Malik et al.~\cite{malik2024flying} to capture a multi-viewpoint flash lidar dataset. 
Unlike Malik et al.~\cite{malik2024flying}, our light source position moves with the camera's viewpoint rather than being stationary with respect to the scene. We capture three scenes: \textit{globe}, \textit{house}, and \textit{spheres}, and  
we use the \textit{statue} scene from Malik et al.~\cite{malik2024flying}, which shows that our method can handle stationary light sources. We provide more details about capture and calibration in the appendix. 

\vspace{-1em}\paragraph{Comparison.} Visually, the captured results follow a similar trend to the simulated ones.
Figure~\ref{fig:capture} shows that we recover more accurate geometry than FWP++, especially for areas where indirect light is present, such as the corners of the walls in the \textit{globe} scene or the bottom of the candles in the \textit{statue} scene.

Table~\ref{tab:results} provides quantitative results for novel view synthesis of lidar measurements. 
While we find that FWP++ shows slight improvements over our approach for view synthesis, we hypothesize that this is because it uses a far less constrained model---our approach is physically grounded and could thus be more sensitive to the calibration of the physical system and model mismatch. 
Note that we omit depth and normal metrics for the captured experiments, as there is no ground truth reference.

\subsection{Additional Results}
\paragraph{Relighting.} Our approach enables time-resolved view synthesis and relighting, which, to our knowledge, has not been demonstrated before from captured multi-view, time-resolved measurements. 
Figure~\ref{fig:teaser} shows relighting results on a novel view from the captured \textit{house} scene with a simulated pulsed source that projects a pattern with the letters ``C-V-P-R''.
In Figure~\ref{fig:relighting}, we show two additional examples of time-resolved relighting: (1) the same \textit{house} scene with three different light sources that converge on the house, and (2) the simulated \textit{kitchen} scene, relit with three point sources that emit pulse trains. 

Note that the indirect component of the radiance cache is predicted using a neural network conditioned on the light source position (Equation~\ref{eqn:tof-cache-indirect}).
So, if the intensity profile of the light source used for relighting differs from the training data (e.g., a projector instead of a uniform point source as in Figure~\ref{fig:teaser}), we use fine-tuning with the radiometric loss $\mathcal{L}_{\textit{dir/indir}}$ of Equation~\ref{eqn:tof-radiometric} and the desired light source profile (additional details are provided in the appendix).

\begin{figure}[t!]
    \begin{center}
    \includegraphics[width=0.48\textwidth]{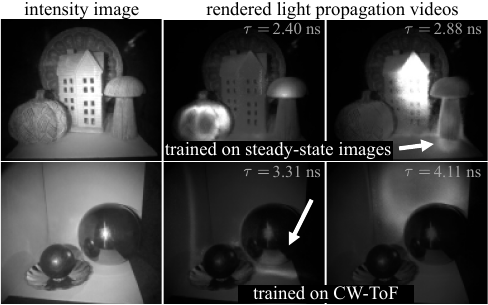}
    \end{center}
        \vspace{-15px}
    \caption{Our approach recovers time-resolved videos of propagating light after training on continuous-wave time-of-flight (CW-ToF) measurements or intensity images. The video frames show direct and indirect light transport effects (arrows).}
    \vspace{-10px}
    \label{fig:sensor_sim}
\end{figure}

\vspace{-1em}
\paragraph{Time-resolved imaging without lidar.}
Although our system recovers time-resolved light transport, it does not necessarily require supervision with lidar measurements. 
Notably, we can train our model using continuous-wave time-of-flight (CW-ToF) measurements or even intensity images, as both can be derived from time-resolved data. We demonstrate this capability by generating time-resolved videos of propagating light based on each input type.
Figure~\ref{fig:sensor_sim} shows the rendered time-resolved measurements after training with either CW-ToF measurements (emulated by convolving the lidar measurements with the CW-ToF illumination waveform) or intensity images (emulated by integrating the lidar measurements over time).
We also demonstrate recovery of direct and indirect light transport effects from the ToF measurements and the steady-state images.
For example, in \textit{house}, we recover diffuse inter-reflections under the mushroom, and in \textit{spheres}, we recover reflections from the ground to the specular sphere (arrows in Figure~\ref{fig:sensor_sim}).

For this application, we only modify the supervision of the model. 
Specifically, we convert the time-resolved predictions to emulate the CW-ToF or steady-state measurements before calculating the loss. After optimization, we directly render the time-resolved video frames.

\vspace{-1em}
\paragraph{Material decomposition.}
Our method also recovers the material parameters for the Disney--GGX BRDF (albedo, roughness, and metalness). 
We visualize these parameters for synthetic and captured scenes in the appendix. 

%% file: sec/discussion.tex
\vspace{-0.5em}
\section{Discussion}
\paragraph{Limitations.} As noted in the results section, our method relies on a more constrained physical model than other approaches, including the FWP++ baseline. 
As such, we notice some performance degradation compared to baselines on captured data, where model mismatch is a potential issue. 
This problem could perhaps be mitigated through better calibration of the physical setup. 
Additionally, our method requires more than one day of optimization on a single GPU due to the time-consuming physical light transport simulation, I/O penalties from loading large time-resolved measurement vectors, and GPU memory bandwidth requirements. 
To address this, it may be possible to use a different neural representation that does not predict the entire time-resolved vector, but instead predicts and supervises the signal at a single time instant. 
Using faster neural representations, like 3D Gaussian Ray Tracing~\cite{3dgrt2024} or EVER~\cite{mai2023neural}, is also an interesting direction.

\vspace{-1em} \paragraph{Impact and future applications.}  Although we do not tackle the problem of non-line-of-sight imaging in this work, it is in principle possible to extend our framework for this application in unconstrained conditions, such as non-line-of-sight imaging with non-planar relay surfaces~\cite{lindell2019wave}. 
Due to the widespread use of lidar technologies, we believe our work has potential for impact in areas such as autonomous navigation or remote sensing --- especially in scenarios with strong indirect lighting effects.

%% file: sec/acknowledgements.tex
\vspace{-1em}
\paragraph{Acknowledgments.}
DBL acknowledges support from NSERC under the RGPIN program, the Canada Foundation for Innovation, and the Ontario Research Fund. BA is supported by a Meta Research PhD Fellowship. MO acknowledges support from NSF CAREER 2238485.

%% file: sec/appendix.tex
\section{Appendix Overview}

We provide additional implementation details related to the architecture of our model, optimization procedure, and experimental settings.
We also include supplemental results and details about the captured dataset.
Code and data are available from our project webpage.
\textbf{Please refer to the webpage and video} for animated visualizations of results, including lidar view synthesis, reconstructed geometry, time-resolved relighting, and separation of direct and indirect light.

\section{Implementation Details}

\subsection{Architecture Details}

\paragraph{Geometry.} We use Zip-NeRF's~\cite{barron2023zip} proposal sampling architecture to represent scene geometry and for volume rendering.
Specifically, we use two hash-encoding-based ``proposal'' networks that output density, which is used for hierarchical sampling, and one final network that outputs the density used in Equation 4, as well as normals $\normal$ used for the cache and physically-based rendering.
The hash-encoding based proposal networks have spatial resolutions of 512 and 1024 along all axes, while the final density network has a resolution of 2048.
Each network has a multi-layer perception (MLP) head with 2 layers and 64 hidden units. 

We use 64 samples for the first proposal network, 64 samples for the second proposal network, and 32 samples for the final geometry network to volume render the cache geometry. 
In order to render the physically-based model, we leverage a single sample quadrature estimator for both primary and secondary rays, as in Attal et al.~\cite{attal2024flash}.

\paragraph{Cache.} The position-dependent appearance feature $\mathbf{f}^{\textit{app}}$ used for the cache has dimension 128 and is predicted with a hash encoding that has a spatial resolution of 2048.
The learned BRDF for the direct component of the cache $f^{\textit{dir}}$ in Equation 11 is a sum of diffuse BRDF $f^{\textit{dir,diff}}(\featureapp)$, and specular BRDF $f^{\textit{dir,spec}}\lft(\featureapp, \normal, \direction_{\light}, \dirout'\rgt)$.
We predict the diffuse BRDF as a function of $\featureapp$ alone, with a 2-layer, 64-hidden-unit MLP.
We predict the specular BRDF as a function of $\featureapp$ as well as the dot product between the normal $\normal$ and normalized half vector $\frac{\direction_{\light} + \dirout'}{|| \direction_{\light} + \dirout' ||}$ with a 2-layer, 64-hidden-unit MLP.

The specular indirect component of the cache, as described in Equation 12, uses a split-sum approximation. We predict $\brdfind(\featureapp, \mathbf{n}, \dirout')$ as a function of the appearance feature and the dot product between normals $\mathbf{n}$ and outgoing direction $\dirout'$. We predict $\radianceinind(\featureapp, \xpos_{\light}, \normal, \dirout')$ as a function of the appearance feature, the reflected direction $\textit{reflect}(\dirout', \mathbf{n})$, and the light source position $ \xpos_{\light}$. Again, both use 2-layer, 64 hidden unit MLPs. We also predict a purely diffuse indirect component $\radianceout^{\textit{indir}, \textit{diff}}$ that is a function of the appearance feature and is conditioned on light source position, with a 2-layer 64-hidden-unit MLP.

\paragraph{Materials.} We leverage the Disney--GGX~\cite{burley2012physically} BRDF parameterization, with parameters albedo $\mathbf{a}(\xpos)$, metalness $m(\xpos)$, and roughness $r(\xpos)$. This BRDF can be written as:
\begin{gather}
    f(\dirin, \dirout, \xpos) = f_{\textit{diffuse}}(\xpos) + f_{\textit{specular}}(\dirin, \dirout, \xpos) \\
    f_{\textit{diffuse}}(\xpos) = \frac{(1 - m(\xpos)) \mathbf{a}(\xpos)}{\pi} \\
    f_{\textit{specular}}(\dirin, \dirout, \xpos) = \frac{DFG}{4 (\normal \cdot \dirin) (\normal \cdot \dirout)}
\end{gather}
\noindent We refer to Burley~\cite{burley2012physically} and Liu \etal~\cite{liu2023nero} for definitions of $(D, F, G)$. We use the Trowbridge-Reitz distribution function~\cite{pharr2023physically} for the normal distribution function $D$.

We predict a material feature $\mathbf{f}_{mat}$ using a hash-encoding-based network with a resolution of 2048, and decode all of the above parameters using a linear layer from this feature.

\paragraph{Importance sampling.} We leverage multiple importance sampling (MIS)~\cite{pharr2023physically}, using the distribution function of the GGX BRDF, and a learned von Mises-Fisher-based importance sampler with an architecture similar to that of Attal et al.~\cite{attal2024flash}. We supervise the importance sampler using the integrated intensity along secondary rays.

\begin{figure}[t!]
    \begin{center}
    \includegraphics[width=\columnwidth]{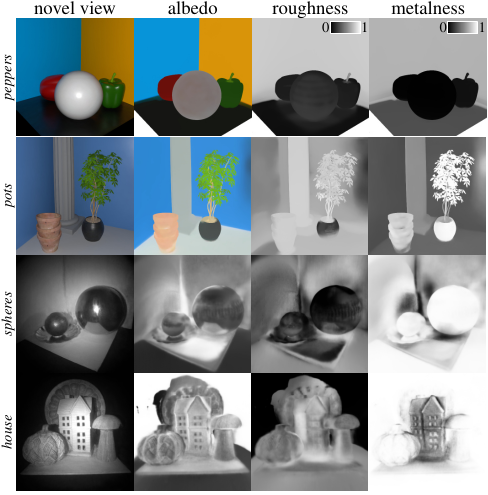}
    \end{center}
        \vspace{-18px}
    \caption{Rendered views and materials for simulated (rows 1--2) and captured scenes (rows 3--4). See the text for a detailed description.}
    \label{fig:materials}
\end{figure}

\subsection{Loss Details}
\paragraph{Mask loss.} The mask loss for a particular ray is defined as:

\begin{align}
    \mathcal{L}_{\textit{mask}} = \textit{mask} \cdot \lft| 1 - \textit{acc} \rgt| + \lft(1 - \textit{mask} \rgt) \cdot \lft| \textit{acc} \rgt|,
\end{align}

\noindent where \textit{acc} is the accumulated transmittance (or sum of the render weights) along a particular ray.

\paragraph{Predicted normal loss.} As discussed, we output the predicted normals using the density hash-encoding-based network. Similar to Ref-NeRF~\cite{verbin2022refnerf} and TensoIR~\cite{jin2023tensoir}, we constrain the predicted normals to match the negative gradient of the density field with an L2 loss:
\begin{align}
    \mathcal{L}_{\textit{normals}} = \sum_{k} w_k \lft\lVert \normal_k^{\textit{pred}} - \normal_k^{\textit{derived}} \rgt\rVert ^ 2,
\end{align}

\noindent where $w_k$ are the render weights for a given ray, and
\begin{align}
    \normal_k^{\textit{derived}} = -\frac{\nabla \sigma(\xpos_k)}{\lVert \nabla \sigma(\xpos_k)\rVert}.
\end{align}
The loss weight $\lambda_{\textit{normals}}$ varies per-dataset. 

\paragraph{Material smoothness loss.} For the smoothness los $\mathcal{L}_{\textit{mat}}$, we leverage the implementation of TensoIR~\cite{jin2023tensoir} for synthetic datasets, and a standard L2 smoothness loss for captured datasets.

\paragraph{RawNeRF loss.} For the photometric losses (Equations 13 and 14 of the main paper), we use $\beta = 1$ for synthetic scenes, $\beta = 2$ for the cache in captured scenes, and $\beta = 1$ for the physically-based model in captured scenes.

\paragraph{Other loss hyperparameters.} For our photometric losses, we set $\lambda_{\textit{cache}} = 10$, $\lambda_{\textit{dir}} = 1$, $\lambda_{\textit{indir}} = 1$.
For the additional losses, we set $\lambda_{\textit{interlevel}} = 0.01$ for all scenes.
For the simulated scenes, we set $\lambda_{\textit{geom}} = 0.0008$, $\lambda_{\textit{disortion}} = 0.0001$, and $\lambda_{\textit{mask}} = 0.1$.
For the captured scenes, we set $\mathcal{L}_{\textit{geom}} = 0.00025$ and $\lambda_{\textit{disortion}} = 0.001$.
We assume that the scene mask is all ones (i.e., all opaque) for captured scenes, and we set the mask loss to $\lambda_{\textit{mask}} = 0.001$.

\subsection{Time-Resolved Imaging Without Lidar}

Section 5.3 of the paper discusses how our model can recover time-resolved videos of propagating light by training on indirect time-of-flight or intensity images. 
In both cases, we write the loss as
\begin{align}
    \mathcal{L}_{\textit{data}} =  \sum_{\xorigin, \dirout} \alpha(\radiancein^{\textit{cache}})  \sum_{k} \Big|\Big| \sum_{\tau} g_k(\tau) \lft(\radiancein - \radianceinmeas \rgt) \Big|\Big|^2.
    \label{eqn:loss-without-lidar}
\end{align}
Here, $\{g_k(\cdot)\}_k$ defines a set of path length importance functions induced by the indirect time-of-flight or intensity sensor~\cite{attal2021torf}. For indirect time-of-flight, we have:
\begin{align}
    g_k(\tau) = \cos(2 \pi f_k \cdot \tau + \theta_k) + 1,
\end{align}
where $f_k$ are frequencies and $\theta_k$ are phase shifts. We use $(f_1, \theta_1) = (30 \times 10^6, 0)$, $(f_2, \theta_2) = (30 \times 10^6, \pi)$, $(f_3, \theta_3) = (170 \times 10^6, 0)$, $(f_4, \theta_4) = (170 \times 10^6, \pi)$. For intensity images, we use $g_1 = 1$.
We apply the same consistency loss as in Equation~15 of the main paper without adjustments.

\subsection{Finetuning for Relighting}

As discussed in Section 5.3 of the paper, we leverage finetuning for relighting whenever the intensity profile of the light source differs from the training data (e.g. a projector as in Fig. 1 of the paper). In order to do this, we freeze all model parameters, apart from those that define the cache direct and indirect appearance (Equation 11 and Equation 12). We then train these parameters in order to minimize the radiometric prior (Equation 15).

\section{Additional Results}

\subsection{Material Decomposition}

In Fig.~\ref{fig:materials}, we show the recovered albedo, roughness, and metalness for simulated and captured scenes from a novel view. 
Qualitatively, the results align with expectations in several respects. The recovered albedo factors out variations in shading and illumination; the roughness is low/dark for specular objects (floor, ball, peppers in row 1; pot in row 2; chrome balls in row 3); and the metalness is bright/high for the pot in row 2 and chrome balls in row 3. 
Generally, the materials are harder to interpret for the captured results---though we expect that improvements to the system calibration would likely improve the results.

We note that for Fig.~\ref{fig:materials}, we leverage an additional loss applied to the \textit{integrated} time-resolved measurements --- specifically the loss in Equation~\ref{eqn:loss-without-lidar} for intensity images. We find that this slightly improves the convergence of the recovered materials. 

\subsection{Additional Baselines}
We include another T-NeRF~\cite{malik2023transient} baseline, which applies a matched filter to the time-resolved measurement to find the direct peak---similar to a conventional lidar---before supervision. We include this result in Table~\ref{tab:supp_add_bas} (see T-NeRF w/ filtering). 

The baseline improves upon T-NeRF and even outperforms FWP++ for geometry modeling. This is expected since one of the main reasons T-NeRF fails in geometry recovery is the presence of the indirect component of light in the lidar scans. However, our method still outperforms this new baseline since the matched filter does not always accurately localize the time of the direct surface reflection, especially under strong indirect light. 
The new baseline also struggles with novel view synthesis since it does not model indirect light transport effects.

\begin{table}
    \captionof{table}{Evaluation of lidar rendering from novel viewpoints and geometry recovery.}
    \label{tab:supp_add_bas}
    \centering
    \setlength{\tabcolsep}{2pt}
    \resizebox{\columnwidth}{!}{
    \begin{tabular}{llcccccc}
        \toprule
        & method & PSNR (dB)$\,\uparrow$ & LPIPS$\,\downarrow$ & SSIM$\,\uparrow$  & MAE$\,\downarrow$  & L1 depth$\,\downarrow$ & T-IOU$\,\uparrow$ \\\midrule
        \parbox[t]{6mm}{\multirow{3}{*}{\rotatebox[origin=c]{90}{\textit{sim}}}} & 
    T-NeRF~\cite{malik2023transient} &  22.44 &  0.40 &  0.71 &  28.00 &  0.59 &  0.58 \\
    & T-NeRF w/ filtering &  \cellcolor{tabthird}24.52 &  \cellcolor{tabthird}0.34 &  \cellcolor{tabthird}0.78 &  \cellcolor{tabsecond}22.54 &  \cellcolor{tabsecond}0.40 &  \cellcolor{tabthird}0.70 \\
    & FWP++~\cite{malik2024flying}    & \cellcolor{tabsecond}29.00 &  \cellcolor{tabfirst}0.30 & \cellcolor{tabsecond}0.87 & \cellcolor{tabthird}22.80 & \cellcolor{tabthird}0.47 & \cellcolor{tabsecond}0.73 \\
    & ours   &  \cellcolor{tabfirst}30.99 & \cellcolor{tabsecond}0.31 &  \cellcolor{tabfirst}0.89 &  \cellcolor{tabfirst}8.45 &  \cellcolor{tabfirst}0.21 &  \cellcolor{tabfirst}0.76
        \\\bottomrule
    \end{tabular}}
\vspace{-1em}
\end{table}

\subsection{Quantitative Results}
We provide a per-scene breakdown of quantitative results for simulated scenes in Table~\ref{tab:results-simulated} and captured scenes in Table~\ref{tab:results-captured}. 
We see similar trends for all scenes as described in the main text.

\subsection{Qualitative Results}
We provide additional qualitative results on novel views in Figure~\ref{fig:supp_simulated} and in the supplemental web page, which includes novel view flythroughs, time-resolved relighting, and separation of direct and indirect light. 
We emphasize that our method recovers more accurate geometry, particularly in scenarios involving strong indirect lighting from specular reflections or diffuse inter-reflections, outperforming previous approaches.

\begin{table*}
    \captionof{table}{Breakdown of results on the simulated scenes for PSNR, LPIPS, SSIM, MAE, L1 Depth (L1) and Transient IOU (T-IOU).}
    \label{tab:results-simulated}
    \centering
    \resizebox{\columnwidth}{!}{
    \begin{tabular}{llcccc}
        \toprule
        & & Pots & Cornell & Peppers & Kitchen \\\midrule

                \parbox[t]{6mm}{\multirow{3}{*}{\rotatebox[origin=c]{90}{\textit{PSNR}}}} & 

     T-NeRF &  \cellcolor{tabthird}23.78 &  \cellcolor{tabthird}23.90 &  \cellcolor{tabthird}19.07 & \cellcolor{tabsecond}23.00 \\
&FWP    & \cellcolor{tabsecond}28.64 & \cellcolor{tabsecond}31.75 & \cellcolor{tabsecond}33.01 &  \cellcolor{tabthird}22.61 \\
&ours   &  \cellcolor{tabfirst}30.44 &  \cellcolor{tabfirst}32.38 &  \cellcolor{tabfirst}37.46 &  \cellcolor{tabfirst}23.68
        \\\midrule

      \parbox[t]{6mm}{\multirow{3}{*}{\rotatebox[origin=c]{90}{\textit{LPIPS}}}} & 
         T-NeRF &  \cellcolor{tabthird}0.36 &  \cellcolor{tabthird}0.32 &  \cellcolor{tabthird}0.44 &  \cellcolor{tabthird}0.49 \\
        & FWP    &  \cellcolor{tabfirst}0.26 &  \cellcolor{tabfirst}0.30 &  \cellcolor{tabfirst}0.26 & \cellcolor{tabsecond}0.39 \\
        & ours   & \cellcolor{tabsecond}0.35 & \cellcolor{tabsecond}0.31 & \cellcolor{tabsecond}0.27 &  \cellcolor{tabfirst}0.30
                \\\midrule
      \parbox[t]{6mm}{\multirow{3}{*}{\rotatebox[origin=c]{90}{\textit{SSIM}}}} & 
        T-NeRF &  \cellcolor{tabthird}0.73 &  \cellcolor{tabthird}0.82 &  \cellcolor{tabthird}0.72 &  \cellcolor{tabthird}0.56 \\
        & FWP    & \cellcolor{tabsecond}0.86 & \cellcolor{tabsecond}0.87 &  \cellcolor{tabfirst}0.94 & \cellcolor{tabsecond}0.79 \\
        & ours   &  \cellcolor{tabfirst}0.90 &  \cellcolor{tabfirst}0.89 & \cellcolor{tabsecond}0.93 &  \cellcolor{tabfirst}0.84
                \\\midrule
          \parbox[t]{6mm}{\multirow{3}{*}{\rotatebox[origin=c]{90}{\textit{MAE}}}} & 
        T-NeRF & \cellcolor{tabsecond}36.09 &  \cellcolor{tabthird}18.33 &  \cellcolor{tabthird}13.03 &  \cellcolor{tabthird}44.56 \\
        & FWP    &  \cellcolor{tabthird}37.41 & \cellcolor{tabsecond}10.86 & \cellcolor{tabsecond}7.20 & \cellcolor{tabsecond}35.75 \\
        & ours   &  \cellcolor{tabfirst}7.81 &  \cellcolor{tabfirst}10.25 &  \cellcolor{tabfirst}2.65 &  \cellcolor{tabfirst}13.08
        \\\midrule
          \parbox[t]{6mm}{\multirow{3}{*}{\rotatebox[origin=c]{90}{\textit{L1}}}} & 
        T-NeRF & \cellcolor{tabsecond}0.18 & \cellcolor{tabsecond}0.10 &  \cellcolor{tabthird}0.42 &  \cellcolor{tabthird}1.66 \\
        & FWP    &  \cellcolor{tabthird}0.29 & \cellcolor{tabsecond}0.10 & \cellcolor{tabsecond}0.28 & \cellcolor{tabsecond}1.20 \\
        & ours   &  \cellcolor{tabfirst}0.04 &  \cellcolor{tabfirst}0.09 &  \cellcolor{tabfirst}0.19 &  \cellcolor{tabfirst}0.53
                \\\midrule
          \parbox[t]{6mm}{\multirow{3}{*}{\rotatebox[origin=c]{90}{\textit{T-IOU}}}} & T-NeRF &  \cellcolor{tabthird}0.66 &  \cellcolor{tabthird}0.69 &  \cellcolor{tabthird}0.76 &  \cellcolor{tabthird}0.20 \\
            & FWP    & \cellcolor{tabsecond}0.82 &  \cellcolor{tabfirst}0.82 & \cellcolor{tabsecond}0.88 & \cellcolor{tabsecond}0.41 \\
            & ours   &  \cellcolor{tabfirst}0.88 & \cellcolor{tabsecond}0.78 &  \cellcolor{tabfirst}0.94 &  \cellcolor{tabfirst}0.46
        \\\bottomrule
    \end{tabular}}
\vspace{-1em}
\end{table*}

\begin{table*}
    \captionof{table}{Breakdown of results on the captured scenes for PSNR, LPIPS, SSIM, MAE and Transient IOU (T-IOU).}
    \label{tab:results-captured}
    \centering
    \resizebox{\columnwidth}{!}{
    \begin{tabular}{llcccc}
        \toprule
        & & House & Globe & Spheres & Statue \\\midrule

                \parbox[t]{6mm}{\multirow{3}{*}{\rotatebox[origin=c]{90}{\textit{PSNR}}}} & 

        T-NeRF &  \cellcolor{tabthird}15.94 &  \cellcolor{tabthird}11.44 &  \cellcolor{tabthird}13.25 &  \cellcolor{tabthird}18.05 \\
        & FWP    & \cellcolor{tabsecond}27.40 &  \cellcolor{tabfirst}26.00 &  \cellcolor{tabfirst}28.51 &  \cellcolor{tabfirst}31.89 \\
        & ours   &  \cellcolor{tabfirst}27.47 & \cellcolor{tabsecond}25.97 & \cellcolor{tabsecond}26.07 & \cellcolor{tabsecond}30.04
        \\\midrule

      \parbox[t]{6mm}{\multirow{3}{*}{\rotatebox[origin=c]{90}{\textit{LPIPS}}}} & 
        T-NeRF &  \cellcolor{tabthird}0.46 & \cellcolor{tabsecond}0.56 &  \cellcolor{tabthird}0.53 &  \cellcolor{tabthird}0.58 \\
        & FWP    &  \cellcolor{tabfirst}0.30 &  \cellcolor{tabfirst}0.34 &  \cellcolor{tabfirst}0.38 &  \cellcolor{tabfirst}0.26 \\
        & ours   & \cellcolor{tabsecond}0.32 &  \cellcolor{tabfirst}0.34 & \cellcolor{tabsecond}0.39 & \cellcolor{tabsecond}0.28
        \\\midrule
      \parbox[t]{6mm}{\multirow{3}{*}{\rotatebox[origin=c]{90}{\textit{SSIM}}}} & 
        T-NeRF &  \cellcolor{tabthird}0.36 & \cellcolor{tabsecond}0.19 &  \cellcolor{tabthird}0.35 &  \cellcolor{tabthird}0.51 \\
        & FWP    & \cellcolor{tabsecond}0.78 &  \cellcolor{tabfirst}0.75 &  \cellcolor{tabfirst}0.81 &  \cellcolor{tabfirst}0.92 \\
        & ours   &  \cellcolor{tabfirst}0.79 &  \cellcolor{tabfirst}0.75 & \cellcolor{tabsecond}0.75 & \cellcolor{tabsecond}0.90                
        \\\midrule
          \parbox[t]{6mm}{\multirow{3}{*}{\rotatebox[origin=c]{90}{\textit{T-IOU}}}} & T-NeRF &  \cellcolor{tabthird}0.34 &  \cellcolor{tabthird}0.10 &  \cellcolor{tabthird}0.13 & \cellcolor{tabsecond}0.34 \\
            & FWP    &  \cellcolor{tabfirst}0.62 &  \cellcolor{tabfirst}0.54 & \cellcolor{tabsecond}0.43 &  \cellcolor{tabfirst}0.60 \\
            & ours   & \cellcolor{tabsecond}0.60 & \cellcolor{tabsecond}0.53 &  \cellcolor{tabfirst}0.44 &  \cellcolor{tabfirst}0.60
            
        \\\bottomrule
    \end{tabular}}
\vspace{-1em}
\end{table*}

\begin{figure*}[h!]
    \vspace{8px}
    \begin{center}
    \includegraphics[width=0.63\textwidth]{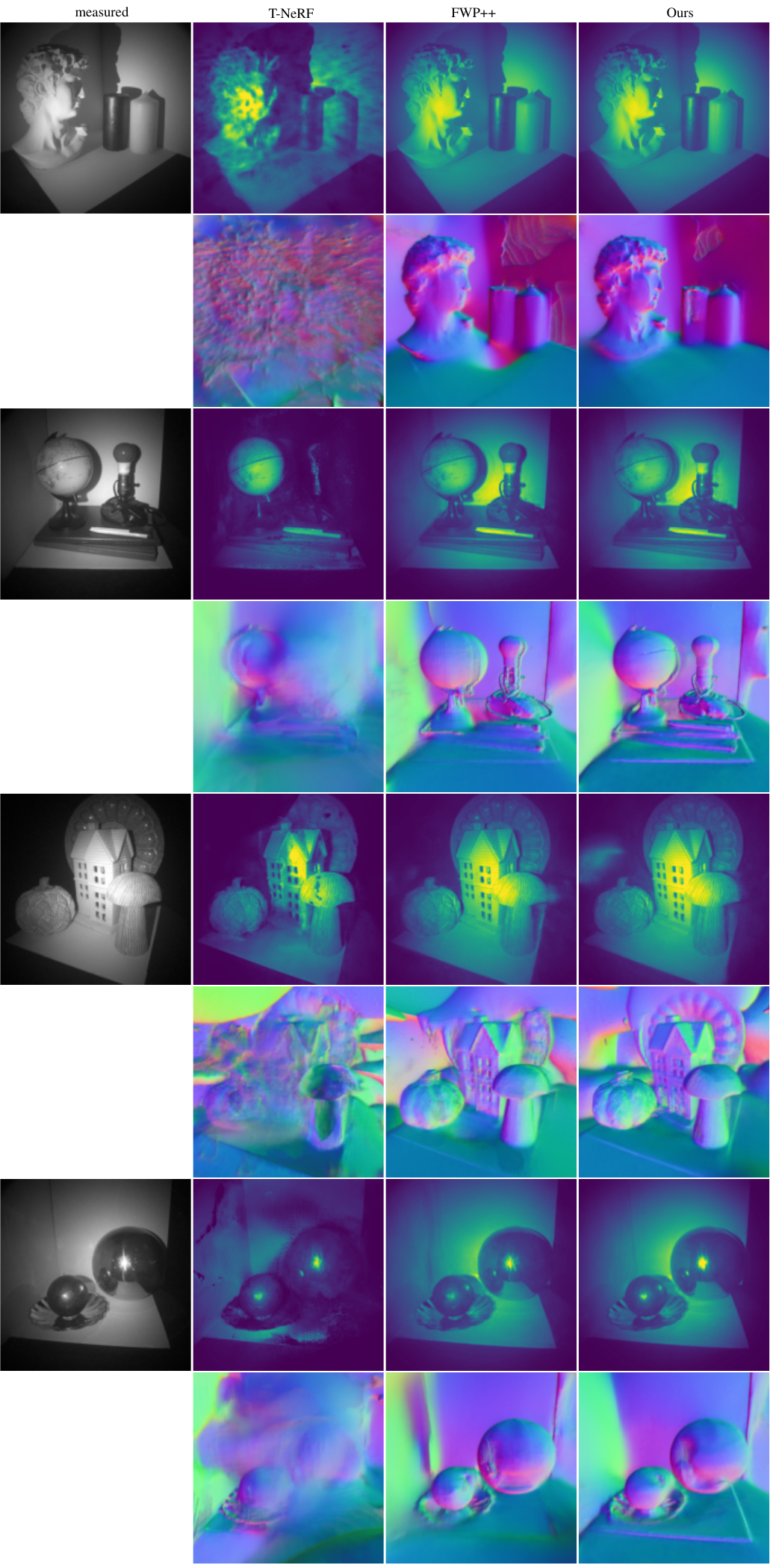}
    \end{center}
        \vspace{-2em}
        \caption{Additional captured results comparing reconstructed normals from the proposed method to those of T-NeRF~\cite{malik2023transient} and FWP++~\cite{malik2024flying}.}
    \vspace{-10px}
    \label{fig:supp_captured}
\end{figure*}

\begin{figure*}[h!]
    \vspace{8px}
    \begin{center}
    \includegraphics[width=0.63\textwidth]{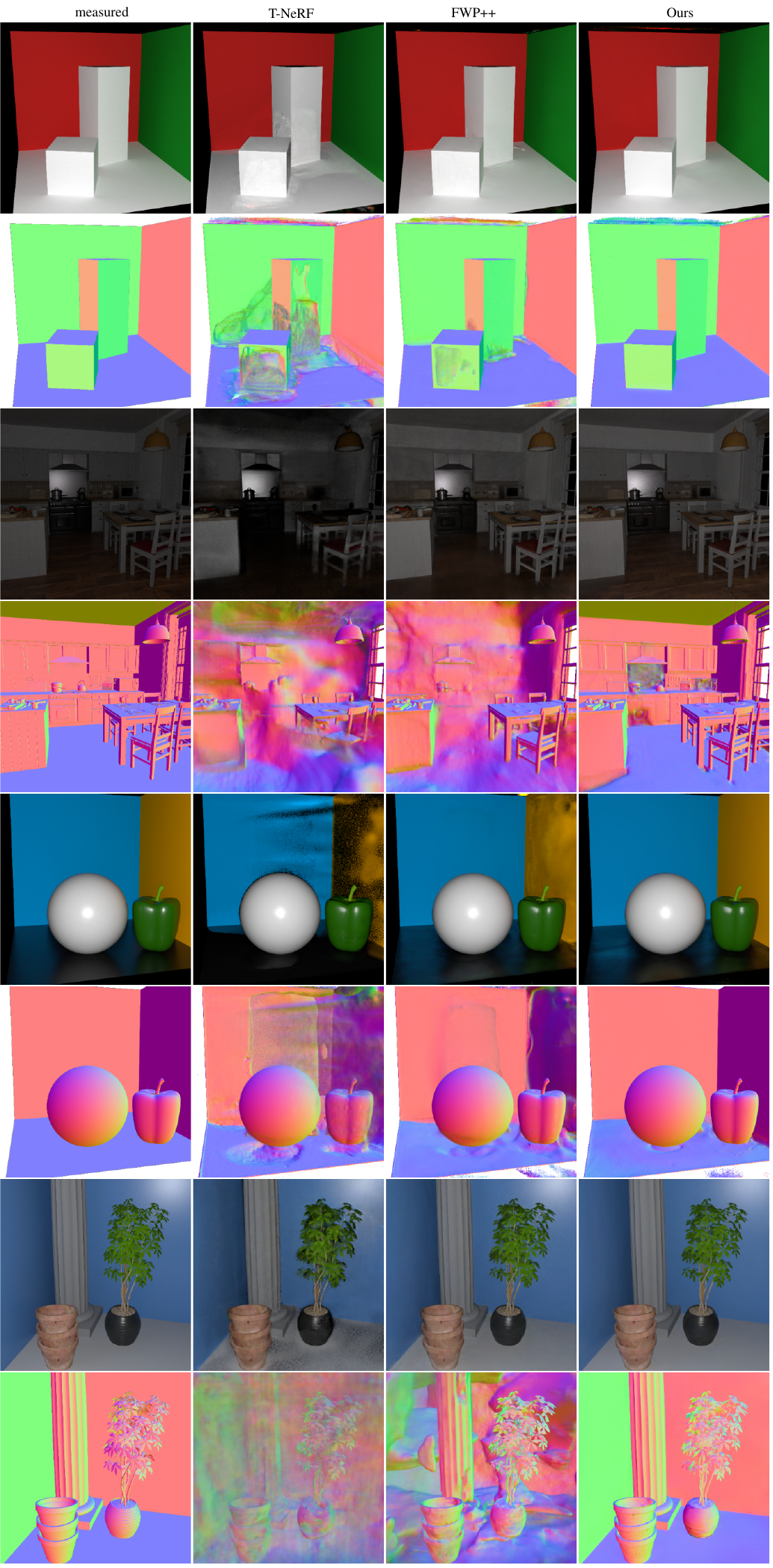}
    \end{center}
        \vspace{-2em}
        \caption{Additional simulated results comparing rendered novel views and reconstructed normals from the proposed method to those of T-NeRF~\cite{malik2023transient} and FWP++~\cite{malik2024flying}.}
    \vspace{-10px}
    \label{fig:supp_simulated}
\end{figure*}

\section{Dataset}
\subsection{Calibration}
To capture our real multi-viewpoint dataset, we use a hardware setup similar to the one used by Malik et al.~\cite{malik2024flying}, with a 532 nm laser emitting 35 ps pulses at a 10 MHz synced with a single pixel scanning SPAD at 512×512 resolution. We capture multiple viewpoints with the same rotation table and elevation arm setup. Specifically, our light source position is fixed for all viewpoints with respect to the camera rather than to the scene. Camera intrinsics are calibrated with a checkerboard and the MATLAB Camera Calibration Toolbox \cite{matlabcamera}, and extrinsics are calibrated using COLMAP \cite{schonberger2016structure} with a scene including a checkerboard so that radial camera pose translation can be scaled by matching the reconstruction to the board's known geometry. 

For our scenes, we assume our light sources are point sources, calibrated so that their location is known with respect to the scene. We simulate point light sources by passing our free-space laser light, coupled through multi-mode fiber, through a collimating lens, and multiple high-power diffusers. To address any residual imperfections in our non-ideal point source, we image a uniformly reflective, diffuse surface with a pre-calibrated pose, using a checkerboard pattern for alignment. This process enables us to compute a directional intensity profile for the light source, which we model during inverse rendering. 

The light source position is calibrated using the following procedure. We (1) capture a checkerboard and compute corner poses, (2) use the corresponding time-resolved measurement for each corner to measure total ToF and thus distance from the light source to the camera, (3) subtract the calibrated corner pose to camera distance, and (4) trilaterate to locate the unknown light source position. 


\subsection{Scene Descriptions}
We provide a description of each captured scene in Table~\ref{tab:scene_desc}.
\begin{table*}[h]
\centering
\caption{Descriptions of the captured scenes. All scenes have a calibrated bin width of 0.0105 m and span 15 degrees in elevation angle.}
\begin{tabular}{lp{5.5cm}>{\centering\arraybackslash}p{1.2cm}>{\centering\arraybackslash}p{1.2cm}>{\centering\arraybackslash}p{1.2cm}>{\centering\arraybackslash}p{2cm}}
\toprule
\textbf{Scene Description} & \textbf{Description} & \textbf{Training Views} & \textbf{Test Views} & \textbf{Azimuth Span} & \textbf{Normalization Scale}  \\
\midrule
House & A diffused pulsed laser source rotates with the lidar sensor and illuminates a ceramic house, mushroom, and pumpkin with a plate in the background. & 81 & 13 & 240° & 600  \\
\midrule
Globe & A diffused pulsed laser source rotates with the lidar sensor and illuminates a globe and a lightbulb. Our model reconstructs the fine details of the wires of the lightbulb stand. & 55 & 11 & 132° & 600  \\
\midrule
Spheres & A diffused pulsed laser source rotates with the lidar sensor and illuminates two specular spheres. & 56 & 11 & 132° & 600  \\
\midrule
Statue & From the Flying with Photons Dataset \cite{malik2024flying}: a stationary diffused pulsed laser source illuminates a statue of David and two candles from the side. & 60 & 15 & 150° & 600  \\
\bottomrule
\label{tab:scene_desc}
\end{tabular}

\end{table*}